# CLUENER2020: FINE-GRAINED NAMED ENTITY RECOGNITION DATASET AND BENCHMARK FOR CHINESE

**Liang Xu**  **Yu tong**  **Qianqian Dong**  **Yixuan Liao**  **Cong Yu**  **Yin Tian**

**Weitang Liu**  **Lu Li**  **Caiquan Liu**  **Xuanwei Zhang**

CLUE Organization

CLUEbenchmark@163.com

## ABSTRACT

In this paper, we introduce the NER dataset from CLUE organization (CLUENER2020), a well-defined fine-grained dataset for named entity recognition in Chinese. CLUENER2020 contains 10 categories. Apart from common labels like person, organization, and location, it contains more diverse categories. It is more challenging than current other Chinese NER datasets and could better reflect real-world applications. For comparison, we implement several state-of-the-art baselines as sequence labeling tasks and report human performance, as well as its analysis. To facilitate future work on fine-grained NER for Chinese, we release our dataset, baselines, and leader-board. [1]

## 1 Introduction

Named-entity recognition (NER) is a sub-task of information extraction that seeks to locate and classify named entities in text into pre-defined categories such as the names of persons, organizations, locations and so on. It is an important basic technique for application fields like information extraction, question answering systems and syntax analysis as an important step in structured information extraction.

NER systems have been created that use linguistic grammar-based techniques as well as statistical models such as machine learning. Hand-crafted grammar-based systems typically obtain better precision, but at the cost of lower recall and months of work by experienced computational linguists. Statistical NER systems typically require a large amount of manually annotated training data. Semi-supervised approaches have been suggested to avoid part of the annotation effort. Pre-training methods such as BERT [1] and their improved versions [2, 3, 4] have led to significant performance boosts across many natural language understanding(NLU) tasks, including NER and other sequence labeling tasks. However, there is a lack of publicly accessible high-quality fine-grained NER dataset for Chinese. Therefore, we collect and release this fine-grained NER dataset, CLUENER2020, to encourage further explorations on this task and other advanced models.

CLUENER2020 contains 10 different categories, including organization, person name, address, company, government, book, game, movie, position, and scene. It contains 13,436 labeled samples. Concretely, each sample contains two parts, the input raw text, and labeled sequences. The raw text is one or two sentences from a piece of news. The labeled sequences are organized as key-value pairs. Keys are categories and values are entities along with their start and end positions. It is worthy of note that it is possible for one category to have more than one entity in a given example.

This dataset is annotated with more categories and details than other available Chinese datasets. As it is more challenging and difficult to complete this task, the capability to differentiate modern models are much better.

In this paper, we offer:

---

[1] https://github.com/CLUEbenchmark/CLUENER2020

(1) A new fine-grained NER dataset for Chinese which is more challenging and with more categories.

(2) Several strong state-of-the-art baselines and their performance for better researches on this task.

(3) A two-stage human performance to compare with modern models.

(4) A public leader-board [2] to automatic test models for this task.

In the following parts, we give a detailed introduction to the construction of the CLUENER2020, the results of baseline methods and human performance on this dataset.

## 2 Dataset Construction and Task Description

| Dataset | Train | Dev | Test | Avg length | Max length | Classes |
|---|---|---|---|---|---|---|
| CLUENER2020 | 10,748 | 1,343 | 1,345 | 37.4 | 50 | 10 |

Table 1: Attribute of CLUENER2020. *Avg length* (*Max length*) means the average (max) sentence length of the CLUENER2020.

| | |
|---|---|
| **sentence:** | 中国国际数码互动娱乐展览会ChinaJoy于2010年7月29日在上海国际博览中心开幕。 |
| **sentence (en):** | *ChinaJoy, the China international digital interactive entertainment exhibition, opened at the Shanghai international expo center on July 29, 2010.* |
| **label:** | address:上海国际博览中心; organization:中国国际数码互动娱乐展览会ChinaJoy |
| **label(en):** | *address: the Shanghai international expo center; organization: ChinaJoy, China international digital interactive entertainment exhibition.* |
| **sentence:** | 环顾脚下的路易港城区，你可以看到硬币上的中央银行，象征着这个国家的金融系统。 |
| **sentence (en):** | *Looking around the city of port Louis, you can see the central bank on a coin, a symbol of the country's financial system.* |
| **label:** | government:中央银行; scene:路易港城区 |
| **label(en):** | *government: the central bank; scene: the city of port Louis.* |
| **sentence:** | 新世纪周星驰逐渐的"慢工出细活"，其自编自导自演的《少林足球》、《功夫》。 |
| **sentence (en):** | *In the new century, Zhou Xingchi gradually became more and more careful in his work. He wrote, directed and acted in Shaolin Soccer and Kung Fu.* |
| **label:** | movie:《少林足球》,《功夫》; name:周星驰 |
| **label(en):** | *movie: Shaolin Soccer, Kung Fu; name: Zhou Xingchi.* |

Table 2: Examples of CLUENER2020. As you can see from the last row of this table, there are two entities belong to same category in a given example.

CLUENER2020 was created from THUCNews[5], which contains around 740,000 news articles retrieved from Sina News RSS. It has 14 news categories from diverse areas , including finance, stock, education, fashion, sports, games, entertainment, and many others.

We use the distant-supervised method [6] with the help of vocabulary to pre-labeling our dataset, then we check and alter some labels manually. We pre-defined some categories of the entity based on sample data. We randomly sample some articles from THUCNews. Each article is a piece of news that belongs to the news category, which has many sentences. We label each sentence one by one through the whole article and ignore those sentences with no entity or too difficult for people to annotate. After labeling, we get distribution for each category and filter those categories with too few data.

To ensure that our dataset is challenging for modern models, we apply data filter technical to the labeled dataset. We call it a cross-validation and filter method. We first split the labeled dataset into a k-folder. For each folder, we train a small size of the modern model(albert_tiny_zh), which represents low model capability compare to other full-model. We predict other folders using a model trained from the current folder; we apply the same process for each folder; eventually, for each folder, we have k-1 predictions. We remove those samples from the folder that all k-1 predictions are all right, which we think are easy samples for our model. We set k equals to 4. We finally get 13,436 labeled dataset with 10 entity categories. The statistical information for CLUENER2020 can be found in Table 1. As is shown in Figure 1, the entity categories distribution is similar for both the training and validation sets. For your better understanding of our dataset, we also list some examples in Table 2.

---
[2]https://www.CLUEbenchmarks.com/ner.html

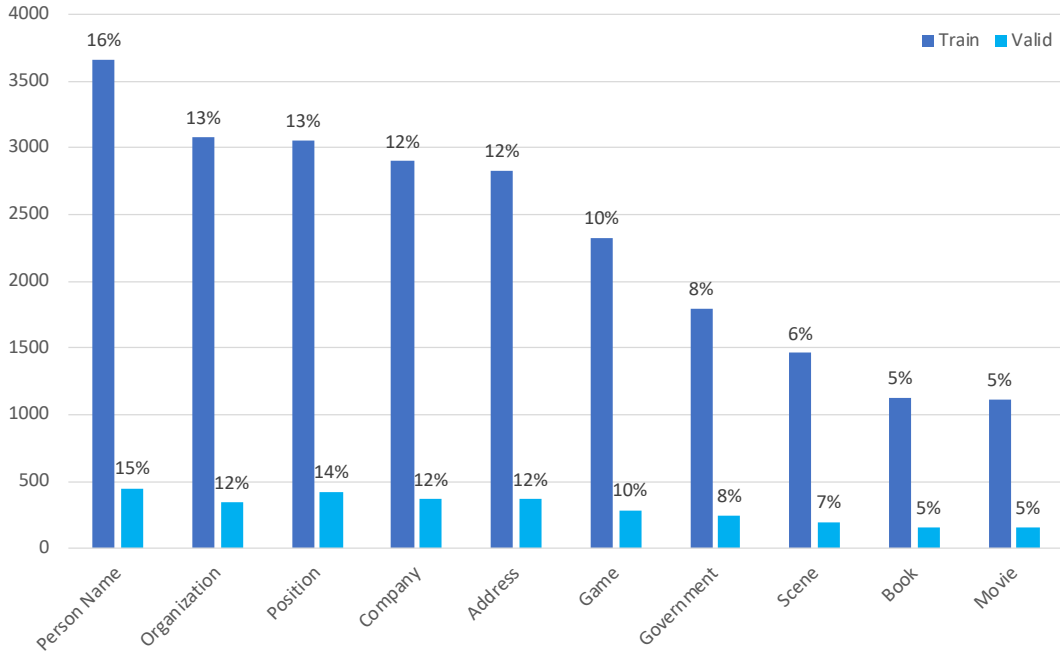

Figure 1: Entities distribution for train and valid set.

## 3 Comparison with Other Chinese NER datasets

We briefly list some information about existing available Chinese NER and include one example for each dataset.

As we can see in Table 3, for MSRANER[7] and PeopleDailyNER[3] dataset, they only have three classic categories (person name, location and organization), while WeiboNER[8, 9] add a category of Geo-political; For BOSONNER[4][10], it add three more categories (time, product name, company name), but the it only has 2k samples.

It should be mentioned that Resume NER [11] owns 8 categories in which *Educational Institution* and *Ethnicity Background* are unique. For Resume NER, the distribution is particularly unbalanced. The category with the largest amount of data is 134 times larger than the category with the smallest amount of data. However, in CLUENER2020, we control the amount of data in each category, making it on the same order of magnitude. See details in Figure 2.

Except those three classic categories, CLUENER2020 has 7 other new categories than MSRANER and PeopleDailyNER, and more samples than BOSONNER. Besides diversity, our dataset is also more challenging than other datasets. Currently, state-of-the-art models in Chinese NER tasks got around f1 score 95 or more, while the best model in CLUENER2020 only got around 80 of the f1 score.

## 4 Experiments

In this section, we implement and evaluate two kinds of typical name entity recognition baselines on CLUENER2020, including traditional systems and pre-trained based models. For pre-trained based baselines, we select BERT and RoBERTa for their effective performance on Chinese tasks currently.

### 4.1 Baselines

To verify the performance of CLUENER2020 on the NER task, we implement the following three baselines systems.

**BILSTM-CRF-NER:** Bidirectional LSTM-CRF models [12] is the most classic method for NER task. Following an embedding layer of input sequences, there are bidirectional LSTM layers to get a contextual representation of an input sequences, then a CRF layer is used to learn some restrictions and rules among entity categories.

---

[3] https://github.com/zjy-ucas/ChineseNER
[4] https://bosonnlp.com/dev/resource

| Dataset | Examples | Entity Categories | Size |
|---|---|---|---|
| **MSRANER** | 据说/o 应/o 老友/o 之/o 邀/o，/o 梁实秋/nr 还/o 坐/o 着/o 滑竿/o 来/o 此/o 品/o 过/o 玉峰/ns 茶/o 。/o <br> (en) It is said that (/o) Liang Shiqiu (/nr) sat on (/o) the pole (/o) to taste(/o) Yufeng(/ns) tea(/o) invited by(/o) old friends(/o). | person name, location, and organization | 50k |
| **PeopleDairyNER** | 19980131-03-001-001/m 克林顿/nr：/w 警告/v 伊拉克/ns 萨达姆/nr：/w 不/d 希望/v 战争/n <br> (en) 19980131-03-001-001 (/m) Clinton (/nr) warns (/v) Iraq's (/ns) Saddam Hussein (/nr) : (/w) we do not (/d) want (/v) war (/n). | person name, location and organization | 23k |
| **BOSONNER** | 此次{{location:中国}}个展，{{person_name:苏珊·菲利普斯}}将与她80多岁高龄的父亲一起合作，哼唱一首古老的{{location:威尔士}}民歌...... <br> (en) In this solo exhibition in {{location:China}}, {{person_name:Susan · Phillips}} will sing an old {{location:welsh}} folk song with her 80-year-old father. | person name, location, organization, time, product name, company name | 2k |
| **WeiboNER** | 我/O 参/O 与/O 了/O 南/B-GPE.NAM 都/I-GPE.NAM 深/B-GPE.NAM 圳/I-GPE.NAM 读/O 本/O 发/O 起/O 的/O 投/O 票/O 涨/O 薪/O 最/O 慢/O 十/O 大/O 行/O 业/O ...... <br> (en) I participated in the poll of the 10 industries with the slowest salary increases launched by South (/B-GPE.NAM) Shenzhen (/I-GPE.NAM) ...... | person name, location, organization, geo-political | 2k |
| **Resume NER** | 美/B-LOC 国/E-LOC 的/o 华/B-PER 莱/I-PER 士/E-PER，我/o 跟/o 他/o 谈/o 笑/o 风/o 生/o ...... <br> (en) Wallace(/B-PER), American(/B-LOC), I(/o) talk(/o) cheerfully(/o) with him(/o) ...... | country, educational institution, location, personal name, organization, profession, ethnicity background and job title | 2k |
| **CLUENER2020** | sentence: 市住建委相关负责人昨天表示，《北京市公共租赁房管理办法》已经获得市委常委会审议原则通过。<br> sentence (en): The charge of municipal commission of housing and urban-rural development said, yesterday, that the measures on the administration of public rental housing in Beijing have been approved by the standing committee of the municipal committee of the communist party of China. <br> label: government: 市住建委, 市常委会; book:《北京市公共租赁房管理办法》; position:负责人 <br> label(en) : government: the municipal commission of housing and urban-rural development, the standing committee of the municipal; book: Measures for the administration of public rental housing in Beijing; position: charge | address, book, company, game, government, movie, name, organization, position and scene | 13K |

Table 3: Comparison with other Chinese NER datasets.

**BERT-NER:** BERT [1] is a Transformer based model that uses pre-training to learn from the raw corpus, and fine-tuning on downstream tasks including NER task.

**RoBERTa-NER:** RoBERTa [3] is an improved version of BERT, which is trained better, longer, and with more data. It removes the next sentence prediction task during the pre-training stage, compared with BERT.

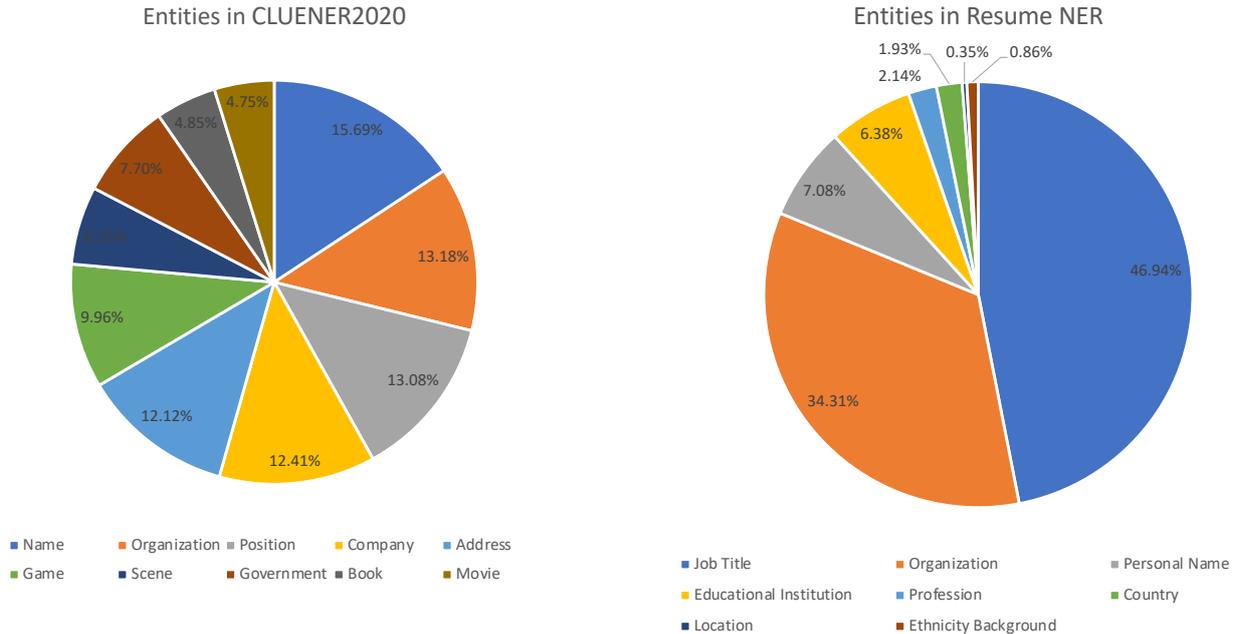

Figure 2: Entities distribution for train set of Resume NER and CLUENER2020.

## 4.2 Implementation Details

We adapt the **BIOS** labeling method to pre-process our dataset. For the **BILSTM-CRF-NER** baseline, the layers of bi-LSTM, the size of hidden states and the size of char embedding are set to 2, 384 and 128 respectively. Training strategies, including dropout and layer normalization, are added to improve the generalization of the model. For the **BERT-NER** baseline, we use the BERT-base pre-trained model. For the **RoBERTa-NER** baseline, we use the RoBERTa-wwm-large [13] pre-trained model. Other specific parameters configuration are shown in Table 4. We train our baselines on training data and select the best model according to the F1 score on the development set. All our models are trained on 1 NVIDIA GPUs. Each of the baseline models predict on the test set, which is submitted to our NER leader-board to get the evaluation results.

| Models | Epoch | Learning Rate | Length | Batch Size |
|---|---|---|---|---|
| BILSTM-CRF-NER | 15 | 1e-2 | 128 | 64 |
| BERT-NER | 4 | 3e-5 | 128 | 32 |
| RoBERTa-NER | 4 | 2e-5 | 128 | 32 |

Table 4: The Experiment Settings for our baseline models.

## 4.3 Human Performance

For better understanding the difficulty of this task and performance of modern models compared with human beings, we conduct an human performance in our experiment.

We follow the "train and predict" two-stage method, just as machine learning models, to conduct human performance. In the training stage, we ask each annotator to be familiar with NER categories and its definitions, then to annotate 30 samples from the development set. They can check ground truth after completing these samples, and they are encouraged to discuss their mistakes and confusion with each other. In the prediction stage, with the knowledge they have learned, annotators are asked to label 100 samples sampled from the test set independently. The final human performance is computed by averaging scores of predictions from three annotators.

From Table 5, Human evaluation is relatively low. We believe there are some cognitive factors. a) human annotators may be unfamiliar with entity definition; b) comparing with models of machine learning, which learn more than 10000 cases from the training set, human annotators only learn 30 cases; c) unlike simple binary classification tasks, this task has 10 different categories; it may difficult for a human. We can see that current state-of-the-art models have strong abilities. The fact that better than the human performance of modern models is quite different from our previous understanding.

### 4.4 Results and Analysis

As seen in Table 5, the performance of pre-training based models, which learned knowledge from the massive amount of raw corpus with self-supervised or transfer learning, are much better than the classic model such as BILSTM+CRF. The best baseline is RoBERTa large model, which outperforms other baseline systems in all entity categories.

However, the f1 score of the best baseline is only around 80, which is much lower than performance on other Chinese NER tasks, such as around f1 score 95 on MSRANER as reported in [14]. It indicates that the NER task with rich categories is still challenging for modern models and leave a big room for improvement.

| Entities | LSTM+CRF | | | BERT | | | RoBERTa | | | Human Performance | | |
|---|---|---|---|---|---|---|---|---|---|---|---|---|
| | P | R | F1 | P | R | F1 | P | R | F1 | P | R | F1 |
| Person Name | 78.06 | 70.42 | 74.04 | 87.93 | 89.58 | 88.75 | 88.01 | 90.21 | **89.09** | 72.34 | 76.77 | 74.49 |
| Organization | 74.04 | 77.97 | 75.96 | 74.89 | 84.56 | 79.43 | 79.34 | 85.57 | **82.34** | 57.41 | 76.00 | 65.41 |
| Position | 71.68 | 68.70 | 70.16 | 75.06 | 83.13 | 78.89 | 78.12 | 81.17 | **79.62** | 59.42 | 51.85 | 55.38 |
| Company | 75.14 | 69.62 | 72.27 | 81.31 | 81.52 | 81.42 | 82.50 | 83.54 | **83.02** | 54.08 | 45.33 | 49.32 |
| Address | 50.00 | 41.75 | 45.50 | 58.70 | 63.25 | 60.89 | 63.27 | 62.00 | **62.63** | 44.10 | 42.03 | 43.04 |
| Game | 87.06 | 83.56 | 85.27 | 85.29 | 87.58 | 86.42 | 85.39 | 88.26 | **86.80** | 84.18 | 76.92 | 80.39 |
| Government | 75.31 | 79.30 | 77.25 | 82.87 | 91.63 | 87.03 | 86.13 | 90.31 | **88.17** | 80.37 | 78.21 | 79.27 |
| Scene | 49.46 | 55.76 | 52.42 | 63.07 | 67.27 | 65.10 | 66.85 | 74.55 | **70.49** | 53.85 | 50.00 | 51.85 |
| Book | 69.42 | 65.12 | 67.20 | 77.12 | 70.54 | 73.68 | 76.42 | 72.87 | **74.60** | 70.71 | 72.73 | 71.70 |
| Movie | 80.45 | 77.54 | 78.97 | 86.13 | 85.51 | 85.82 | 86.52 | 88.41 | **87.46** | 80.95 | 51.85 | 63.21 |
| Overall@Macro | 71.06 | 68.97 | 70.00 | 77.24 | 80.46 | 78.82 | 79.26 | 81.69 | **80.42** | 65.74 | 62.17 | 63.41 |

Table 5: The results on CLUENER2020.

## 5 Conclusion

In this work, we release the fine-grained name entity recognition dataset for Chinese, CLUENER2020. It is more challenging than other existing Chinese datasets. CLUENER2020 reserves more detailed annotations, which is consistent with real-world scenarios. We conduct experiments using our strong baselines; We also compare the performance of models with humans. Experiments demonstrate that fine-grained NER is still challenging and leave plenty of room to make improvements.